\documentclass{iise}

\usepackage{multirow}
\usepackage{hyperref}
\usepackage{url}
\usepackage{fancyhdr}
\usepackage{graphicx}%
\usepackage{multirow}%
\usepackage{amsmath,amssymb,amsfonts}%
\usepackage{amsthm}%
\usepackage{mathrsfs}%
\usepackage[title]{appendix}%
\usepackage{xcolor}%
\usepackage{textcomp}%
\usepackage{manyfoot}%
\usepackage{booktabs}%
\usepackage{algorithm}%
\usepackage{algorithmicx}%
\usepackage{algpseudocode}%
\usepackage{listings}%
\usepackage{array}%
\usepackage{adjustbox}%
\usepackage{colortbl}%
\usepackage{afterpage}%
\usepackage{caption}%
\usepackage{subcaption}%
\usepackage{color}
\usepackage{wrapfig}
\usepackage{float}

\usepackage[most]{tcolorbox} 

\conference{}

\title{\titlesize Multi-Agent LLMs for Adaptive Acquisition in Bayesian Optimization}

\author{
Mohammadsina Almasi, Andrea Carbonati, Hadis Anahideh\footnote{Corresponding Author: Hadis Anahideh (hadis@uic.edu)}\\
University of Illinois Chicago, IL, USA}
\authorlist{Almasi, Carbonati and Anahideh}

\begin{document}
\maketitle

\begin{abstract}
\small{The exploration–exploitation trade-off is central to sequential decision-making and black-box optimization, yet how Large Language Models (LLMs) reason about and manage this trade-off remains poorly understood. Unlike Bayesian Optimization, where exploration and exploitation are explicitly encoded through acquisition functions, LLM-based optimization relies on implicit, prompt-based reasoning over historical evaluations, making search behavior difficult to analyze or control. 
In this work, we present a metric-level study of LLM-mediated search policy learning, studying how LLMs construct and adapt exploration–exploitation strategies under multiple operational definitions of exploration, including informativeness, diversity, and representativeness.
We show that single-agent LLM approaches, which jointly perform strategy selection and candidate generation within a single prompt, suffer from cognitive overload, leading to unstable search dynamics and premature convergence. To address this limitation, we propose a multi-agent framework that decomposes exploration–exploitation control into strategic policy mediation and tactical candidate generation. A strategy agent assigns interpretable weights to multiple search criteria, 
while a generation agent produces candidates conditioned on the resulting search policy defined as weights. This decomposition renders exploration–exploitation decisions explicit, observable, and adjustable. 
Empirical results across various continuous optimization benchmarks indicate that separating strategic control from candidate generation substantially improves the effectiveness of LLM-mediated search.}
\end{abstract}

\section*{Keywords}
Large Language Models, Exploration-Exploitation Trade-Off, Black-Box Optimization

\section{Introduction}\label{sec:intro}

Balancing exploration and exploitation is a fundamental challenge in sequential decision-making and black-box optimization, determining how an algorithm allocates limited evaluation budgets between probing uncertain regions of the search space and refining currently promising solutions~\cite{srinivas2009gaussian}. Classical treatments of this trade-off are mathematically explicit. In Bayesian Optimization (BO), exploration and exploitation are encoded directly through acquisition functions such as Upper Confidence Bound (UCB) and Expected Improvement (EI), where posterior uncertainty and mean estimates jointly guide sampling decisions under well-characterized regret guarantees~\cite{garnett2023bayesian}. 
While this explicit encoding yields theoretical clarity and performance guarantees, it also confines exploration–exploitation behavior to a narrowly prescribed design space. In particular, exploration strategies are fully specified by fixed acquisition function templates and pre-defined uncertainty models, limiting how search behavior can be modified or restructured beyond the assumed probabilistic framework~\cite{almasi2025fairpilot}.

Recent advances in LLMs have introduced a fundamentally different paradigm to sequential decision-making and black-box optimization. Rather than relying on explicit probabilistic surrogates or analytically defined acquisition functions, LLM-based optimizers operate through in-context learning and natural-language reasoning over historical evaluations and candidate descriptions~\cite{yang2023large}. This implicit reasoning enables optimization in settings that are difficult to formalize, such as problems with mixed data types, complex constraints, or semantic objectives, but also obscures how exploration and exploitation are balanced during search. 
Most existing work treats LLMs as opaque heuristics, focusing on end-to-end performance~\cite{liu2024large} without analyzing how exploration notions are prioritized or how search strategies evolve.
This opacity is particularly problematic because exploration itself is not a single concept. In practice, exploration may be defined in terms of informativeness (sampling where uncertainty is high)~\cite{kim2024active}, diversity (covering distinct regions of the search space)~\cite{hong2018diversity}, or representativeness (maintaining coverage of the underlying data distribution)~\cite{du2015exploring}, which are typically encoded explicitly in BO \cite{nezami2025enhancing} but may be implicitly mixed by LLM-based optimizers. 
Understanding how LLMs perceive and operationalize different notions of exploration is therefore critical for assessing their suitability as optimization agents.

In this work, we study LLMs through the lens of LLM-mediated search policy learning. Rather than treating the LLM as a surrogate model or as a black-box optimizer, we view it as a policy mediator that constructs a search strategy by reasoning over problem context, historical evaluations, and candidate exploration criteria. Importantly, this mediation occurs without updating the LLM’s parameters or replacing the underlying surrogate model. Instead, the LLM adaptively composes a search policy, expressed through interpretable weights over exploration and exploitation criteria, that guides candidate selection at each iteration.
We begin by analyzing a single-agent LLM baseline, in which a single prompt is responsible for jointly selecting an exploration–exploitation strategy and generating new candidate points. While conceptually simple, this design places substantial cognitive demands on the LLM. This coupling leads to unstable search behavior in practice.

To address these limitations, we propose a multi-agent framework that decomposes exploration–exploitation control into two complementary roles. A \textit{strategy agent} is responsible for selecting and weighting multiple search criteria, including exploitation, informativeness, diversity, and representativeness, based on the current optimization context. A \textit{generation agent} then produces candidate solutions conditioned on this explicit search policy. This division of labor mirrors the separation between acquisition design and candidate optimization in classical BO, while preserving the flexibility of LLM-based reasoning. Crucially, it renders exploration–exploitation decisions explicit, observable, and interpretable at each iteration.
By tracking the temporal evolution of metric weights across optimization iterations, we provide direct empirical evidence that LLMs can exhibit highly adaptive, context-sensitive search strategies, dynamically reallocating emphasis between exploitation, exploration, and auxiliary criteria in response to the observed landscape. This effect is especially pronounced in engineering problems with heterogeneous, task-dependent sensitivities, where classical Gaussian Process surrogates often yield miscalibrated uncertainty due to restrictive assumptions.
 
Overall, this work makes three contributions:
(1) a metric-level framework for analyzing exploration–exploitation behavior in LLM-driven optimization,
(2) a principled decomposition of LLM-mediated search policy learning into strategic and tactical components, and
(3) an empirical characterization of when and how LLMs succeed, or fail, as BO search policy mediators.

\section{Related Work}\label{sec:lr}

Recent work has investigated LLMs as black-box optimizers, leveraging their in-context learning and capacity to reason over unstructured representations of optimization trajectories~\cite{liu2024language}. OPRO (Optimization by PROmpting) formulates optimization as an iterative, derivative-free process in which natural-language prompts encode historical solution–objective pairs~\cite{yang2023large}. By exposing sorted trajectories, OPRO enables heuristic refinement without explicit surrogates or gradients; however, exploration–exploitation is regulated only implicitly via temperature control. As a result, the approach exhibits erratic search dynamics in continuous spaces, frequently collapsing toward suboptimal regions and displaying pronounced dependence on initial prompt configurations.
These limitations are systematically characterized by Huang et al.~\cite{huang2024exploring}, who evaluate LLMs on discrete and continuous black-box problems and identify deficiencies in numerical precision, scalability, and invariance properties. Most notably, they reveal severe exploration–exploitation pathologies, with models oscillating between near-uniform sampling and greedy collapse as prompt order varies, indicating that LLMs act as heuristic population-level samplers rather than principled optimizers with stable strategic control.

To mitigate these issues, several works embed LLMs within classical Bayesian Optimization (BO) pipelines. LLAMBO translates surrogate modeling and acquisition into structured language reasoning, enabling zero-shot warm-starting and conditional candidate generation under sparse data~\cite{liu2024large}. While sample-efficient in early stages, LLAMBO retains externally fixed exploration hyperparameters and exhibits sensitivity to observation ordering, requiring ad hoc stabilization. BOLT instead distills high-quality BO trajectories into an LLM to produce strong task-specific initializations~\cite{zeng2025large}. Although effective in large-scale multi-task settings, exploration–exploitation behavior is absorbed implicitly through data distillation, remaining opaque and uncontrollable at inference time.
Other hybrid methods combine probabilistic surrogates with LLM-based generation more tightly. BOPRO conducts BO in a latent embedding space, using acquisition-guided retrieval to bias LLM prompts toward promising regions~\cite{agarwal2025searching}. Its performance depends critically on the fidelity of the embedding space and still relies on a single LLM to process retrieved context and generate candidates, leaving strategic behavior implicit and unobservable. Conceptually related, Song et al.~\cite{song2024position} recast black-box optimization as sequence modeling with Transformer-based foundations, emphasizing transferability but offering no architectural mechanism for explicit exploration–exploitation regulation. Similarly, Zhang et al.~\cite{zhang2023using} show strong early-stage performance for LLM-driven hyperparameter optimization in high-dimensional spaces, yet retain a monolithic, single-agent design where strategy and generation are entangled, leading to re-exploration, instability, and limited interpretability.


\section{Methodology}\label{sec:method}

We consider a black-box optimization problem defined over a bounded domain $\mathbf{X} \subset \mathbb{R}^d$ , where the objective function $f: \mathbf{X} \rightarrow \mathbb{R}$ is expensive to evaluate and available only through sequential queries. At iteration $t$, the optimizer has access to a history of evaluated points $\mathcal{D}_{t} = \{(\mathbf{x}_i, y_i)\}_{i=1}^{t}$ and must select a new candidate $\mathbf{x}_{t+1}$ to identify a global maximizer $\mathbf{x}^* = \arg\max_{\mathbf{x} \in \mathbf{X}} f(\mathbf{x})$ under fixed evaluation budget $T_{\max}$. 


\subsection{Exploration and Exploitation Metrics}
To make exploration–exploitation decisions explicit and analyzable, we define a set of interpretable, metric-based criteria that characterize different aspects of search behavior. These metrics are computed externally from the optimization history and candidate set, and are provided to the LLM as inputs rather than learned implicitly. Specifically, we consider the following criteria: i)  exploitation, favoring candidates expected to improve upon the best observed objective (e.g., via normalized improvement scores); ii) informativeness, favoring candidates in under-explored or uncertain regions (approximated by local sparsity or novelty); iii) diversity, encouraging global coverage through dispersion in the search space; and iv) representativeness, promoting candidates that reflect the overall structure of the sampled domain (e.g., proximity to empirical cluster centers).




\subsection{LLM-Mediated Search Policy Learning}

We frame the role of the LLM as search policy mediation. At each iteration, the LLM reasons over the current optimization context and constructs a search policy expressed as a weighted combination of the predefined criteria. Formally, a search policy at iteration $t$ is represented as a weight vector.
Crucially, these weights are not learned through parameter updates or reinforcement learning. Instead, they are produced episodically by the LLM through in-context reasoning over (i) the optimization history $\mathcal{D}_{t}$ (ii) summaries of recent search behavior (e.g., stagnation, improvement trends), and
(iii) problem descriptors such as dimensionality or observed landscape structure.
This formulation makes the exploration–exploitation trade-off explicit, interpretable, and adjustable at each iteration.
\begin{wrapfigure}{r}{0.38\linewidth}
\vspace{-1em}
\begin{tcolorbox}[
  enhanced,
    title={Algorithm 1: Multi-Agent LLM Optimization Loop},
  colback=white,
  colframe=black,
  boxrule=0.6pt,
  left=12pt,
  right=12pt,
  top=12pt,
  bottom=12pt,
  before upper=\raggedright
]
\small
\textbf{Input:} black-box function $f$, budget $T_{\max}$, initial data $\mathcal{D}_0$.
\begin{enumerate}[leftmargin=1.4em, itemsep=2pt]
  \item Construct strategy prompt $P_t^{\text{strat}}$ from $\mathcal{D}_{t-1}$.
  \item Infer metric weights
        $W_t \leftarrow \text{LLM}_{\text{Agent1}}(P_t^{\text{strat}})$.
  \item Construct generation prompt $P_t^{\text{gen}}$ from $\mathcal{D}_{t-1}$ and $W_t$.
  \item Generate candidate
        $\mathbf{x}_t \leftarrow \text{LLM}_{\text{Agent2}}(P_t^{\text{gen}})$.
  \item Evaluate $y_t = f(\mathbf{x}_t)$ and update
        $\mathcal{D}_t \leftarrow \mathcal{D}_{t-1} \cup \{(\mathbf{x}_t, y_t)\}$.
\end{enumerate}

\textbf{Return:} $x^{\text{best}} = \arg\max_{i=1,\dots,T_{\max}} y_i$.

\textbf{Return:} $x^{\text{best}}$
\end{tcolorbox}
\vspace{-2em}
\end{wrapfigure}
\textbf{Single-Agent LLM Baseline.}
As a baseline, we consider a single-agent LLM that jointly performs search policy construction and candidate generation within a single prompt.
Given the optimization history $\mathcal{D}_{t}$ at iteration $t$ and problem description, the LLM is instructed to reason about exploration and exploitation and directly output one or more new candidate points $\mathbf{x}_t$. 
While conceptually simple, this design couples \textit{high-level strategic reasoning} with \textit{low-level tactical decision-making}, two forms of reasoning that differ substantially in nature and cognitive demands. High-level strategic reasoning concerns intent and policy: deciding whether to prioritize exploration or exploitation at a given iteration, diagnosing whether the search has stagnated, selecting which notion of exploration to emphasize (e.g., diversity, uncertainty, or improvement), and determining how aggressively to explore at the current stage of optimization. These decisions are inherently abstract, contextual, and comparative, as they require weighing trade-offs across competing objectives and interpreting the global state of the search. For example, a strategic decision may take the form: ``\textit{The search appears to be stagnating; exploration via increased diversity should be emphasized}.''
In contrast, \textit{low-level tactical decisions} concern execution. These include selecting the exact candidate point ($\mathbf{x}_{t+1} \in \mathbb{R}^d$) to evaluate, specifying its coordinate values, and choosing which candidate most faithfully instantiates the selected strategy. Such decisions are concrete, numerical, and constraint-driven, requiring precision rather than abstraction. By forcing a single LLM prompt to simultaneously reason about strategic intent and produce precise numerical candidates, the single-agent design conflates fundamentally different reasoning tasks, often leading to inconsistent strategy application and unstable search behavior.
\textbf{Multi-Agent Decomposition.}
To address the limitations of the single-agent approach, we propose a modular \textit{Multi-Agent} framework that decomposes search policy mediation into distinct roles, inspired by the separation between acquisition design and candidate optimization in classical BO.
The \textit{strategy agent} is responsible for constructing the search policy. It receives a summarized representation of ($\mathcal{D}_t$), recent optimization progress, and metric definitions, and outputs a weight vector ($w_t$) over the exploration and exploitation criteria. By isolating this task, the strategy agent focuses exclusively on high-level reasoning about search behavior and trade-offs.
The \textit{generation agent} receives the weighted search policy ($w_t$) and produces candidate points by optimizing a scalarized score derived from the weighted criteria. This agent does not reason about strategy selection; instead, it operationalizes the provided policy to generate concrete candidates.

\subsection{Prompt Engineering and Interface}\label{subsec:prompt-eng}

We employ a structured prompt interface to mediate all interactions between the optimization loop and the LLM agents, ensuring consistency, interpretability, and reproducibility. Each agent receives a fixed prompt template with clearly delineated sections for optimization history, metric definitions, and task-specific instructions, reducing sensitivity to prompt wording. We utilize a structured templating system implemented via the LangChain library~\cite{langchain}.

At iteration ($t$), the optimization history ($\mathcal{D}*{t-1} = {(\mathbf{x}_i, y_i)}*{i=1}^{t}$) is serialized in a consistent format and provided to the agents. To manage context length, the history may be summarized using salient statistics; unless otherwise noted, our experiments include the full history within the evaluation budget ($T_{\max}$).

The \textit{strategy agent} receives explicit natural-language definitions of the exploration and exploitation criteria and outputs a normalized vector of metric weights in a predefined structured format. The \textit{generation agent} conditions candidate generation on these weights, treating them as explicit constraints rather than advisory guidance. Outputs from both agents are machine-readable to support reliable integration within the optimization loop.

{Prompt templates, metric definitions, and output schemas are held fixed throughout the experiments to isolate the effects of LLM-mediated search policy reasoning from prompt engineering artifacts.
This interface makes exploration–exploitation decisions explicit, observable, and controllable.
}

\subsection{Prompt Samples}

To facilitate reproducibility and provide transparency, we present representative examples of the prompt templates used in our experiments. These samples illustrate how the theoretical framework described in Section \ref{sec:method} is operationalized into concrete natural language instructions.

\begin{figure}[H]
    \centering
    \includegraphics[width=0.98\textwidth]{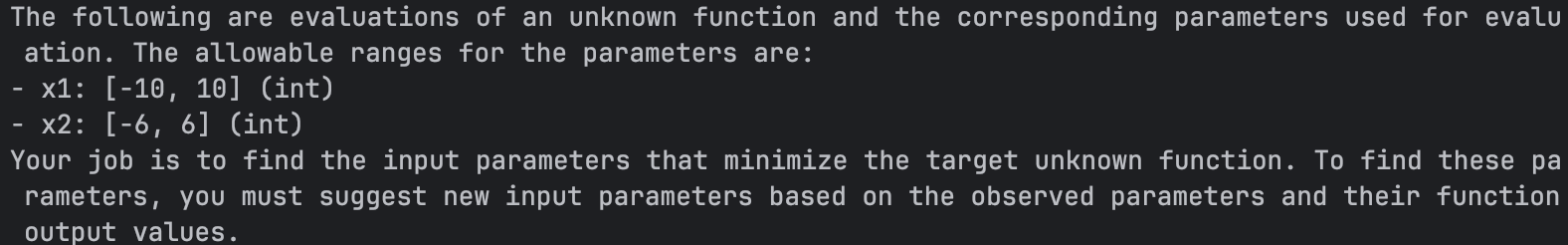}
    
    \caption{Information Encoding Prompt Sample.}
    \label{fig:info_encoding_prompt}
\end{figure}

The foundation of the prompt chain is the context initialization shown in Figure \ref{fig:info_encoding_prompt}. This segment explicitly defines the optimization problem by encoding the domain constraints (e.g., integer ranges for $x_1, x_2$) and the specific optimization objective (minimization). This ensures the agent is grounded in the feasible search space before attempting any reasoning.

\begin{figure}[H]
    \centering
    \begin{subfigure}[b]{0.98\textwidth}
        \centering
        \includegraphics[width=\linewidth]{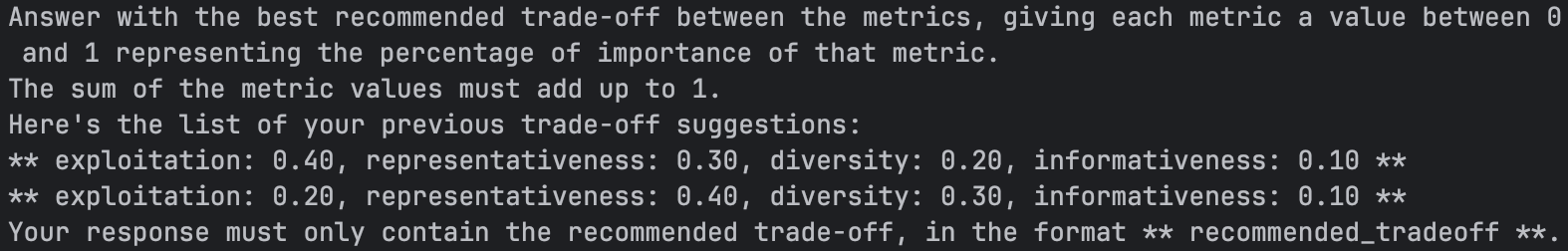}
        \caption{Metric Selection Structured Output Format}
        \label{fig:metrics_output_prompt}
    \end{subfigure}
    
    \begin{subfigure}[b]{0.98\textwidth}
        \centering
        \includegraphics[width=\linewidth]{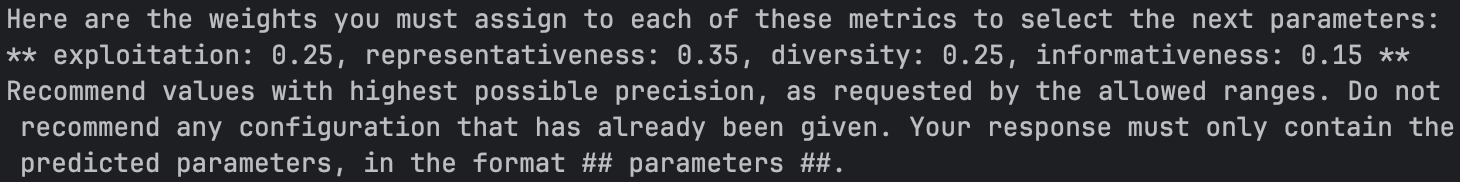}
        \caption{Candidate Generation Structured Output Format}
        \label{fig:candidate_output_prompt}
    \end{subfigure}
    
    \caption{Structured Output Format Prompt Samples.}
    \label{fig:ouptut_format_prompts}
\end{figure}

A critical challenge in LLM-driven engineering is the stochastic nature of text generation. To mitigate parsing failures, we enforce strict output protocols as demonstrated in Figure \ref{fig:ouptut_format_prompts}. Figure \ref{fig:metrics_output_prompt} illustrates the instruction for the Strategist Agent, requiring it to output weight vectors enclosed in specific delimiters (\texttt{** weights **}). Similarly, Figure \ref{fig:candidate_output_prompt} restricts the Generation Agent to output candidate parameters in a machine-readable format (\texttt{\#\# parameters \#\#}), enabling seamless integration with the programmatic evaluation loop.

\begin{figure}[H]
    \centering
    \includegraphics[width=0.98\textwidth]{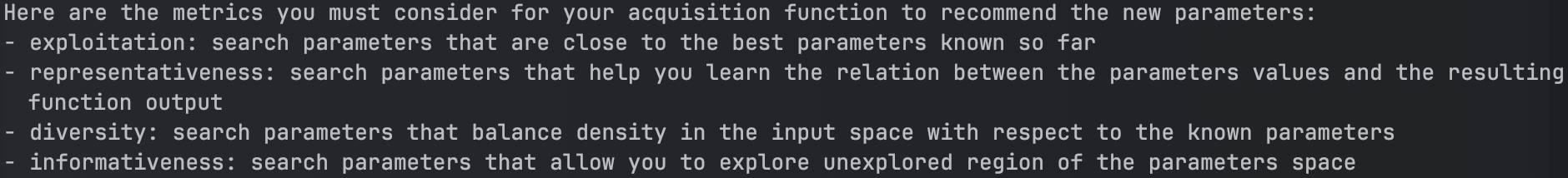}
    
    \caption{Metric Definitions Prompt Sample.}
    \label{fig:metric_defs_prompt}
\end{figure}

Finally, Figure \ref{fig:metric_defs_prompt} displays the semantic definitions provided to the agents. By clearly defining abstract optimization concepts we provide the necessary semantic anchors for the LLM to reason effectively about the trade-offs required at each iteration.

\section{Experiments and Results}\label{sec:experiments}




\textbf{Setup.} To evaluate the proposed multi-agent LLM framework, we conduct experiments across three representative problem classes: a classical analytical benchmark (Rosenbrock, with no semantic or structural information exposed in the prompt)~\cite{anahideh2022high}, a high-dimensional machine learning task (hyperparameter tuning)~\cite{nezami2023hyperparameter}, and a complex engineering control problem (robot pushing)~\cite{astudillo2021functionnetworks}. These domains span varying levels of dimensionality, smoothness, and evaluation cost.
All LLM-based experiments use the \texttt{llama-3.3-70b-instruct} model hosted on UIC-managed infrastructure. Each experiment is conducted under a fixed evaluation budget of 30 function queries, with initial configurations sampled randomly using fixed seeds for reproducibility. 

\subsection{Performance and Trade-off Analysis Across Benchmark Classes}
\begin{figure}[H]
    \centering
    \begin{subfigure}{0.28\textwidth}
        \centering
        \includegraphics[width=\linewidth]{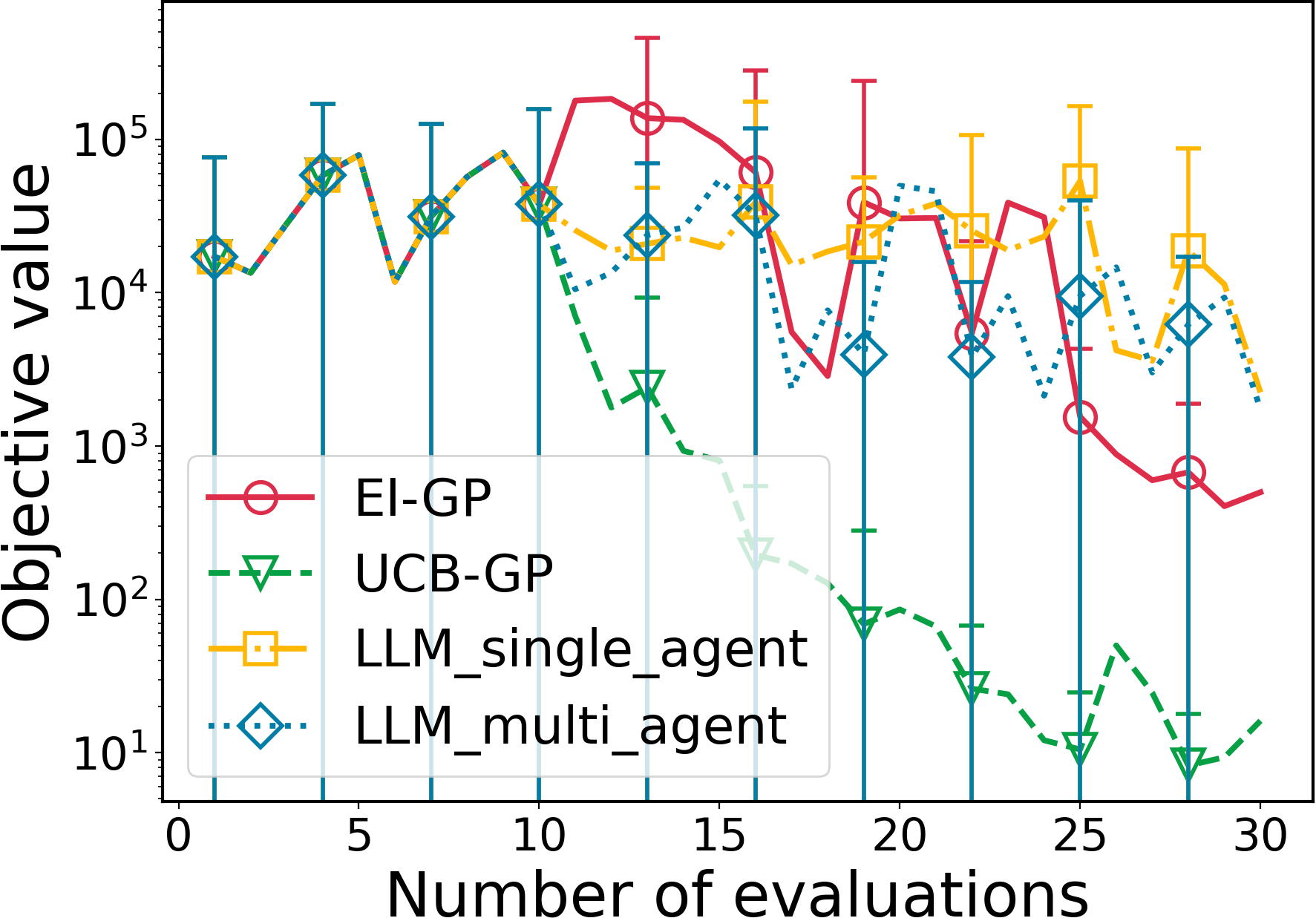}
        \caption{Rosenbrock Performance}
        \label{fig:rosen_multi}
    \end{subfigure}
    \begin{subfigure}{0.28\textwidth}
        \centering
        \includegraphics[width=\linewidth]{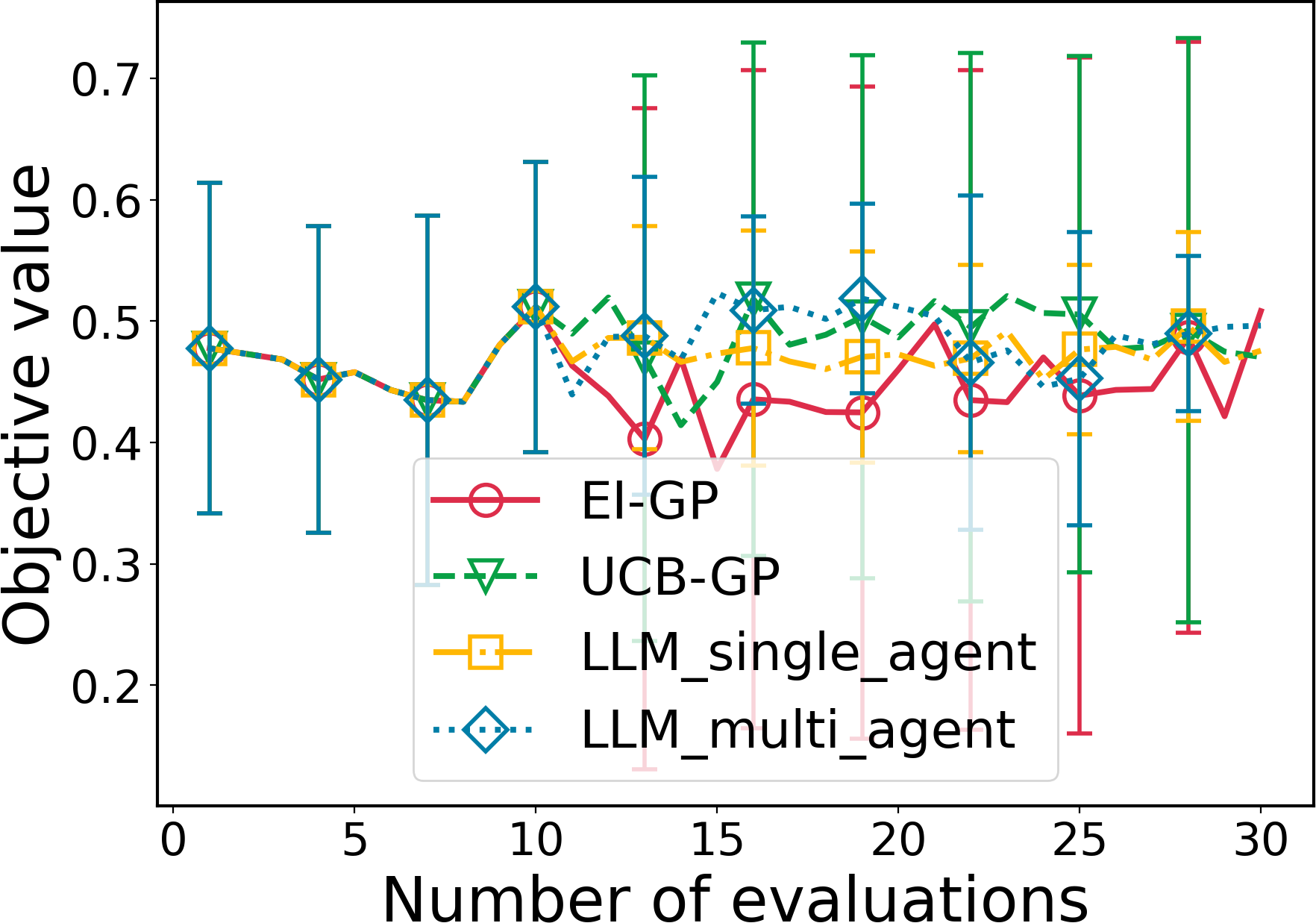}
        \caption{HPT Performance}
        \label{fig:hpt_multi}
    \end{subfigure}
    \begin{subfigure}{0.285\textwidth}
        \centering
        \includegraphics[width=\linewidth]{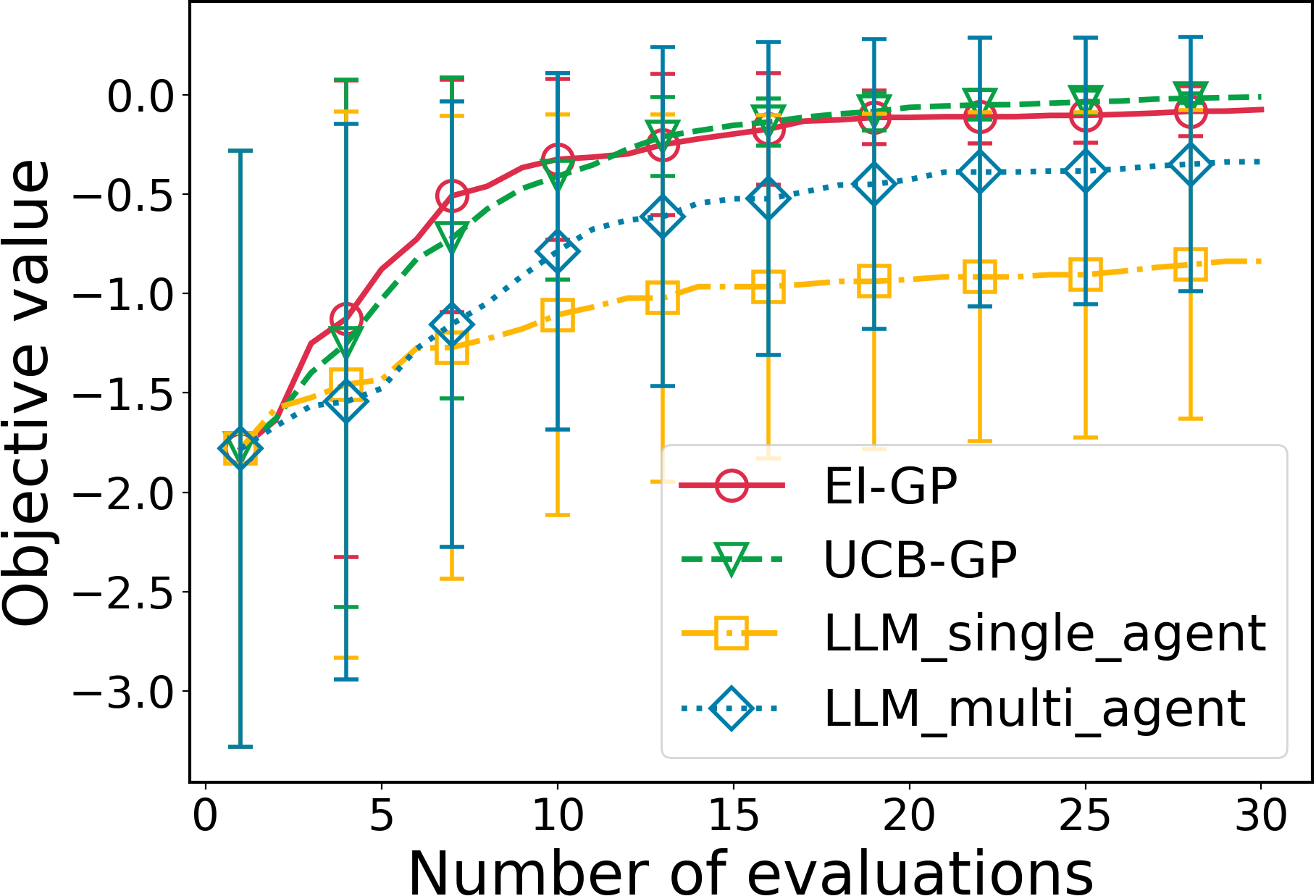}
        \caption{Robot Pushing Performance}
        \label{fig:robot_multi}
    \end{subfigure}
\begin{subfigure}{0.28\textwidth}
    \centering
    \includegraphics[width=\linewidth]{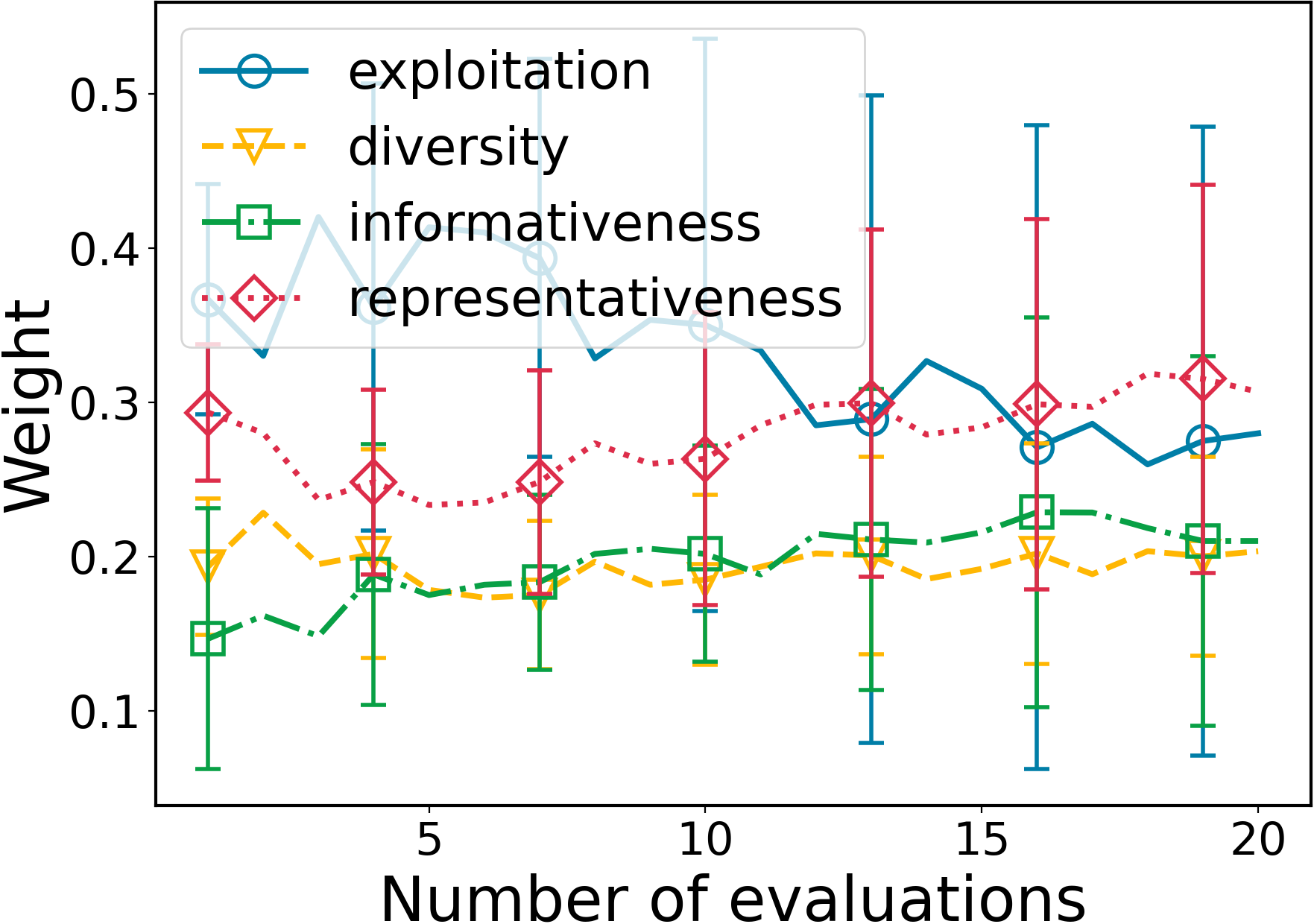}
    \caption{Rosenbrock Metric Weights}
    \label{fig:rosen_multi_metrics}
\end{subfigure}
\begin{subfigure}{0.28\textwidth}
    \centering
    \includegraphics[width=\linewidth]{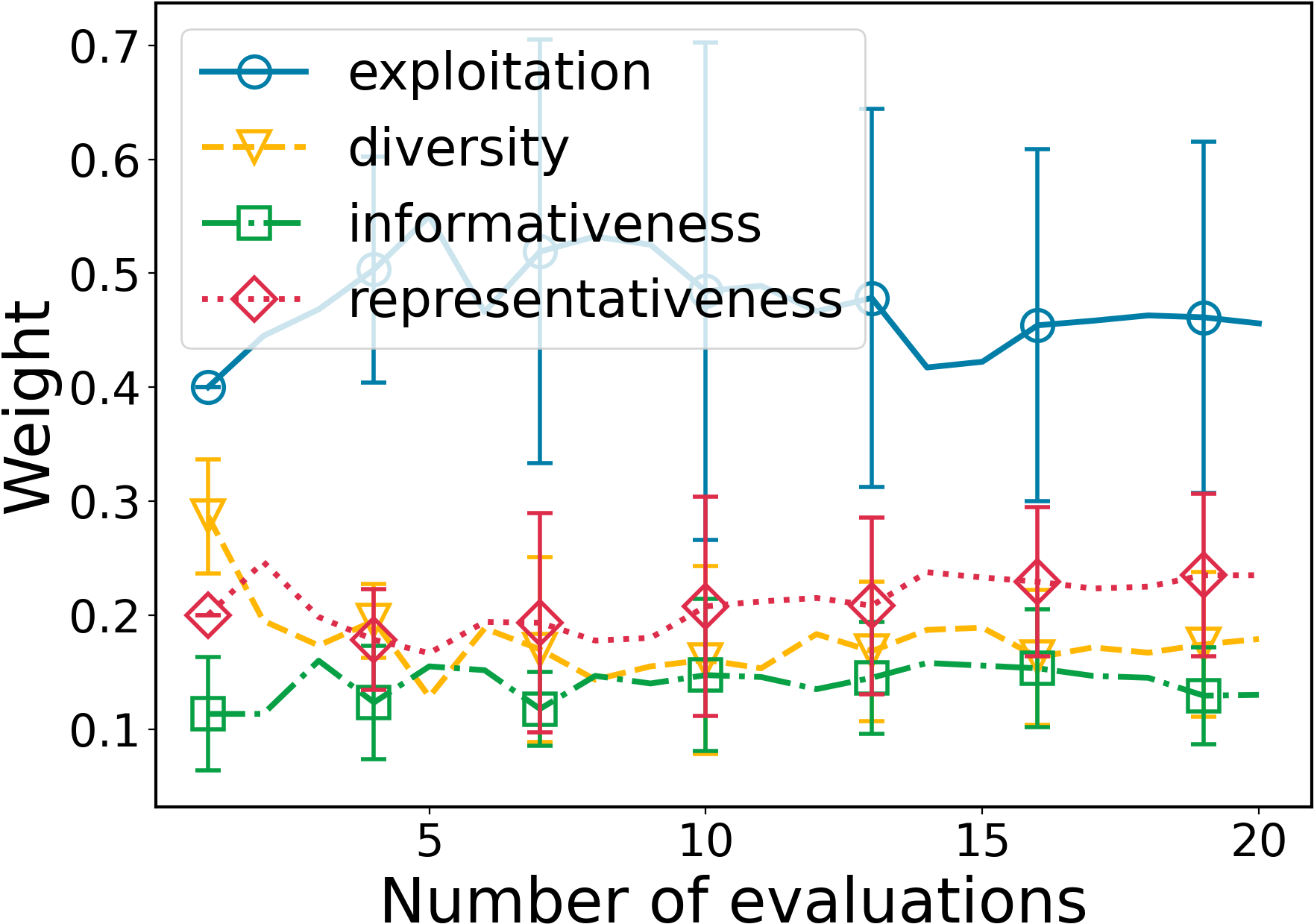}
    \caption{HPT Metric Weights}
    \label{fig:hpt_multi_metrics}
\end{subfigure}
\begin{subfigure}{0.28\textwidth}
    \centering
    \includegraphics[width=\linewidth]{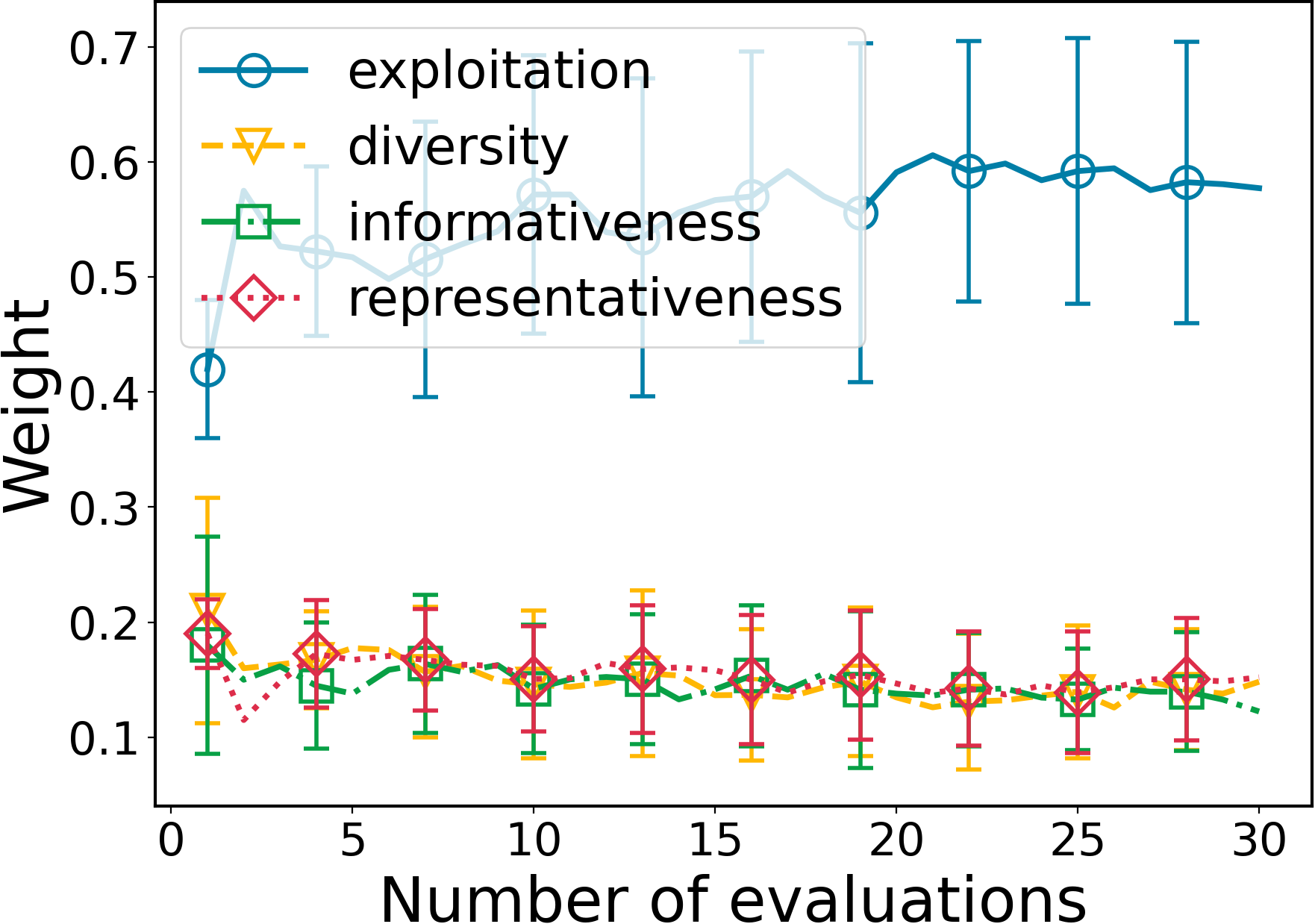}
    \caption{Robot Pushing Metric Weights}
    \label{fig:robot_multi_metrics}
\end{subfigure}

\caption{Performance evaluation of single- and multi-agent approaches.}
\label{fig:exp_results}
\end{figure}

Across benchmarks, the behavior of LLM-based optimizers depends strongly on landscape structure, revealing both their strengths and limitations. On the Rosenbrock function (Figures~\ref{fig:rosen_multi}–\ref{fig:rosen_multi_metrics}), EI and UCB achieve smooth and consistent convergence by effectively exploiting the narrow valley geometry of the objective. On the other hand, LLM-based methods exhibit noisier trajectories with intermittent exploratory jumps and slower convergence, with the single-agent variant particularly unstable. Although the multi-agent approach is more controlled, it still fails to achieve the fine-grained numerical refinement required in this smooth setting. Strategy weight trajectories show an initial emphasis on \textit{exploitation} followed by a shift toward exploration-oriented criteria, indicating that slow but correct progress is interpreted as stagnation—an adaptation that is misaligned with the Rosenbrock landscape and leads to degraded performance.

In the hyperparameter tuning task (Figures~\ref{fig:hpt_multi}–\ref{fig:hpt_multi_metrics}), the performance gap narrows substantially. EI and UCB exhibit higher variance and slower improvement, while the multi-agent LLM achieves competitive, and in some runs superior, performance. The single-agent LLM remains less stable, reflecting difficulty in consistently balancing strategic reasoning and candidate generation. Correspondingly, metric weights show a persistent emphasis on \textit{exploitation}, suggesting a focused refinement strategy that is effective in discrete, high-dimensional search spaces.

\begin{figure}[h!]
    \centering
    \begin{subfigure}{0.3\textwidth}
        \centering
        \includegraphics[width=\linewidth]{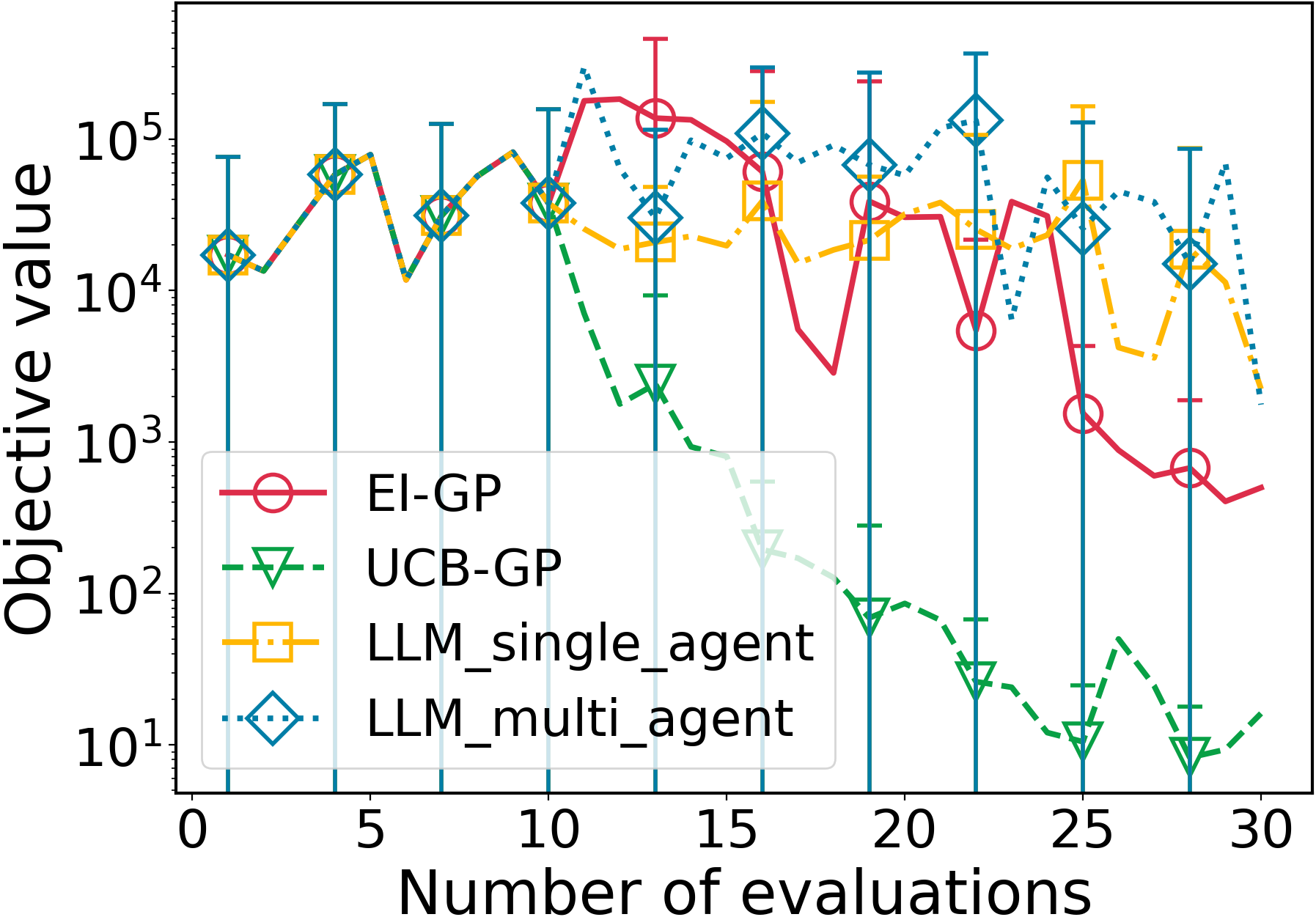}
        \caption{Representativeness Results}
        \label{fig:rosen_repr}
    \end{subfigure}
    \begin{subfigure}{0.3\textwidth}
        \centering
        \includegraphics[width=\linewidth]{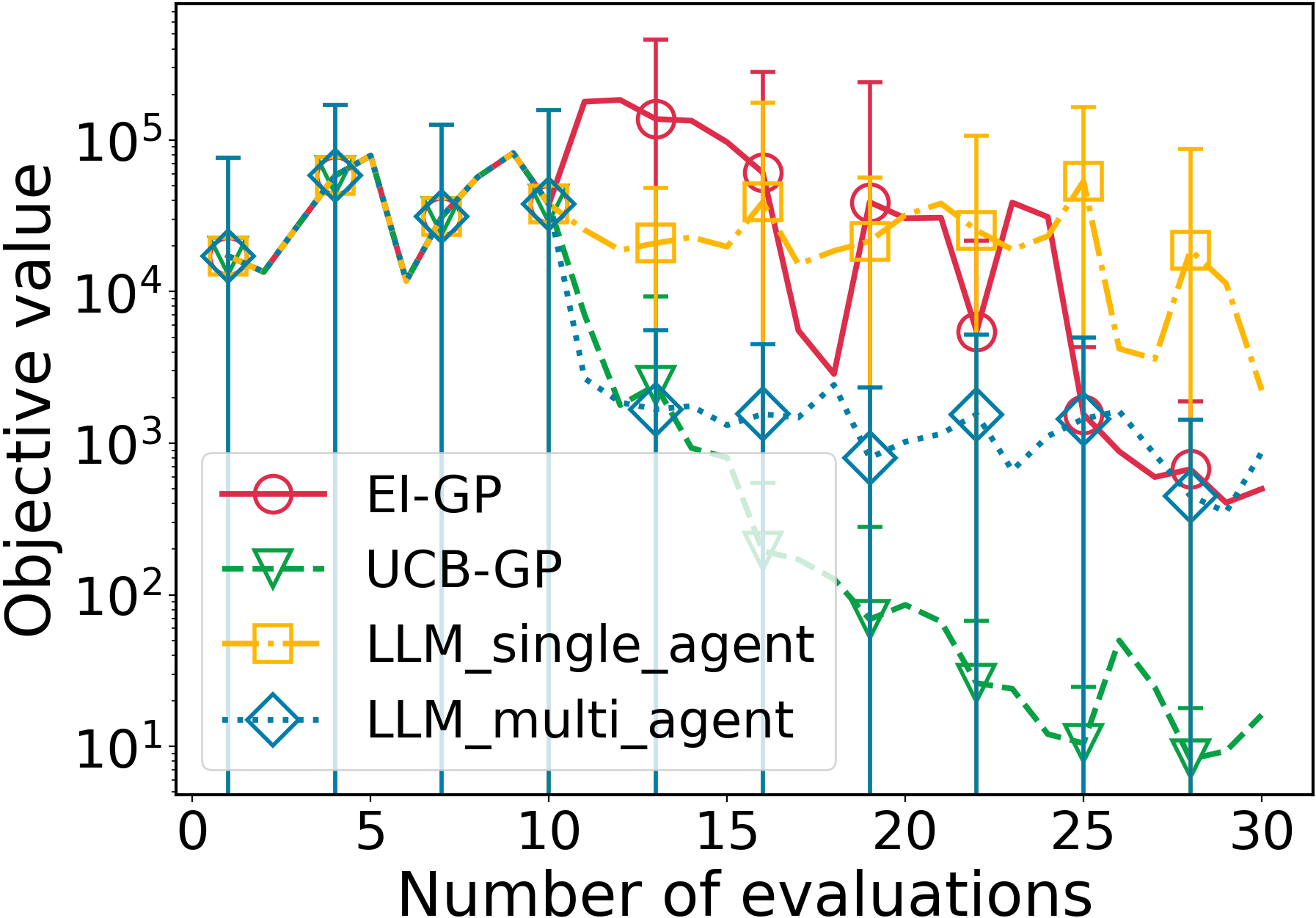}
        \caption{Diversity Results}
        \label{fig:rosen_div}
    \end{subfigure}
    \begin{subfigure}{0.3\textwidth}
        \centering
        \includegraphics[width=\linewidth]{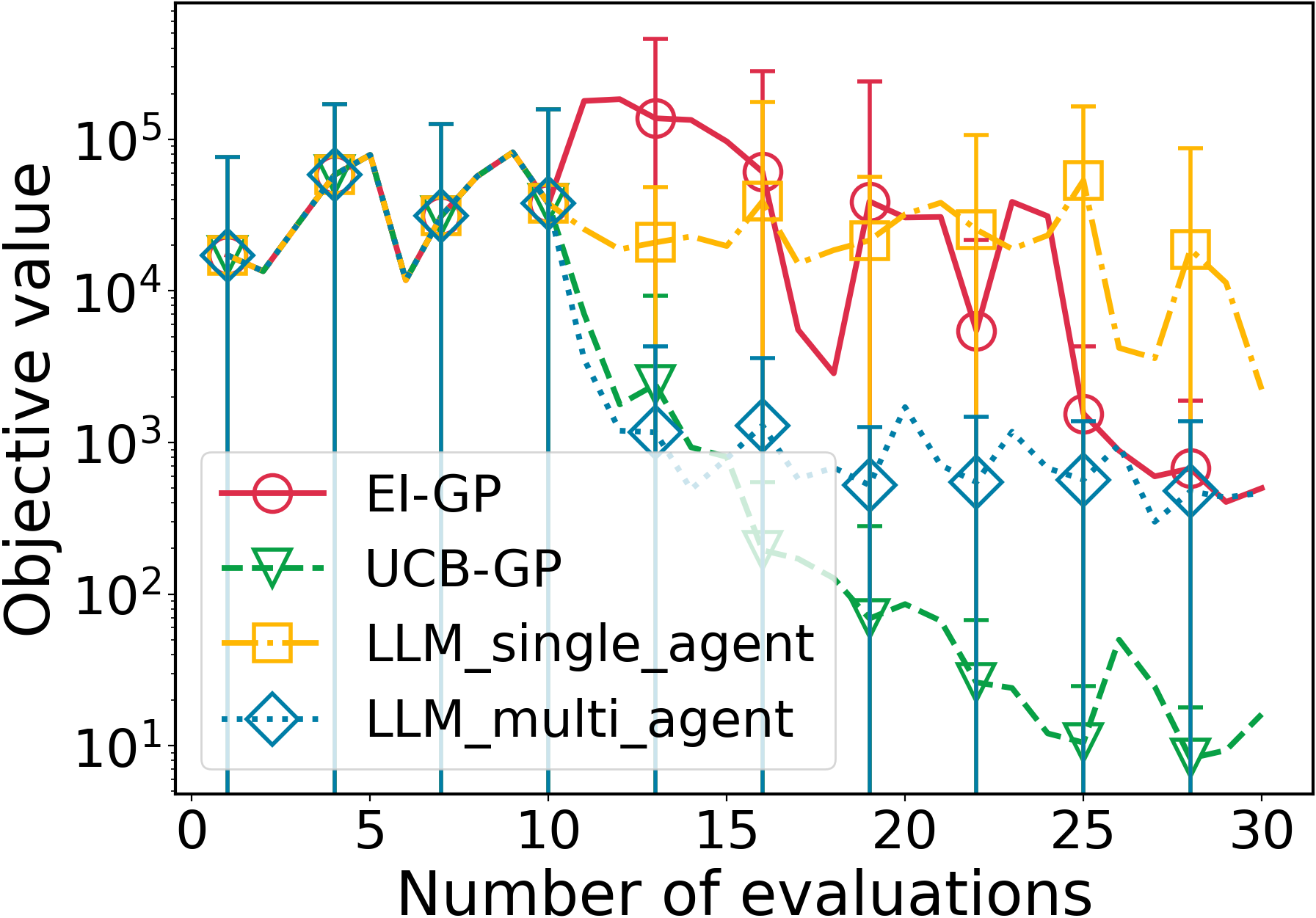}
        \caption{Informativeness Results}
        \label{fig:rosen_info}
    \end{subfigure}

    \begin{subfigure}{0.3\textwidth}
        \centering
        \includegraphics[width=\linewidth]{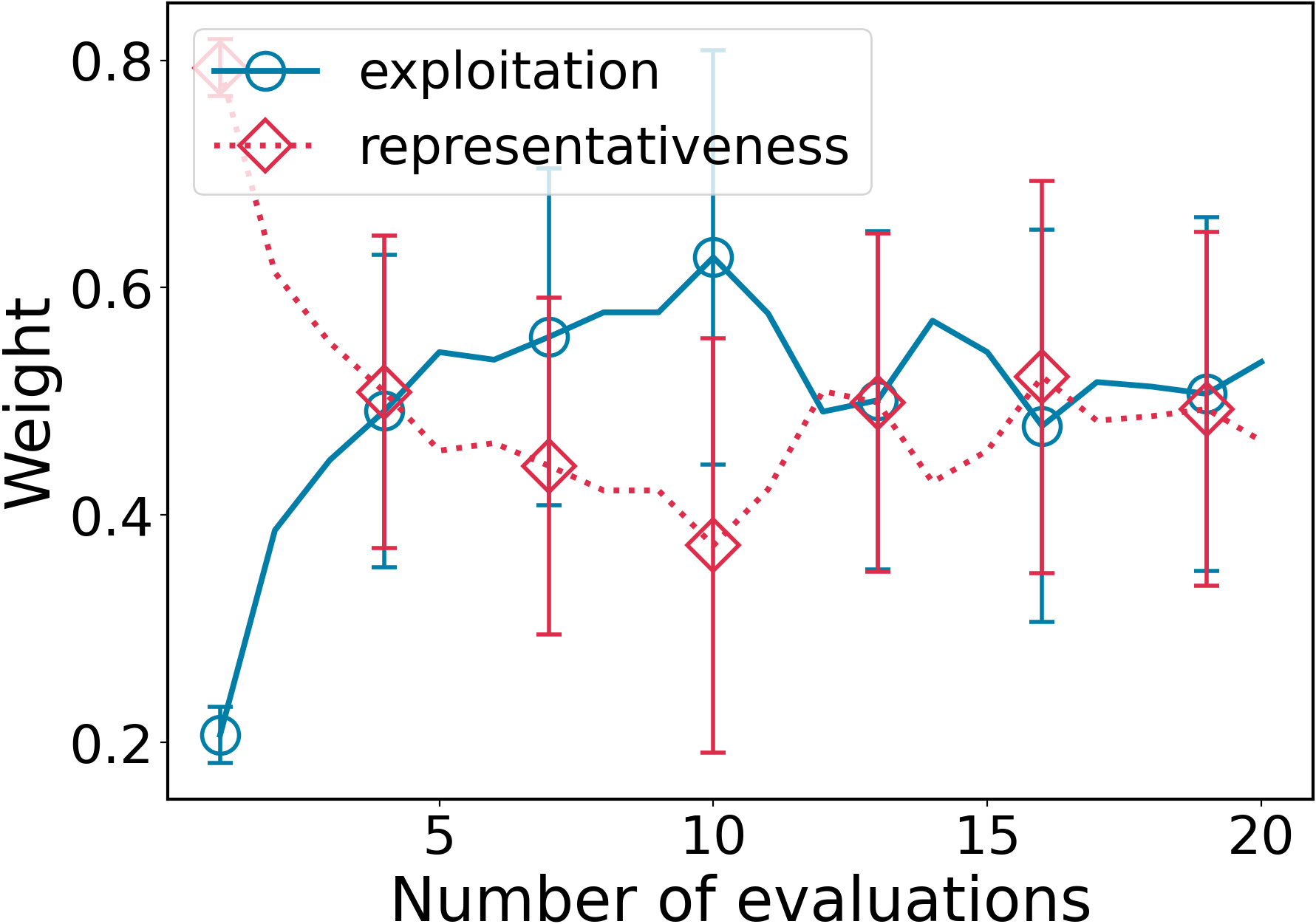}
        \caption{Representativeness Metrics}
        \label{fig:rosen_repr_metrics}
    \end{subfigure}
    \begin{subfigure}{0.3\textwidth}
        \centering
        \includegraphics[width=\linewidth]{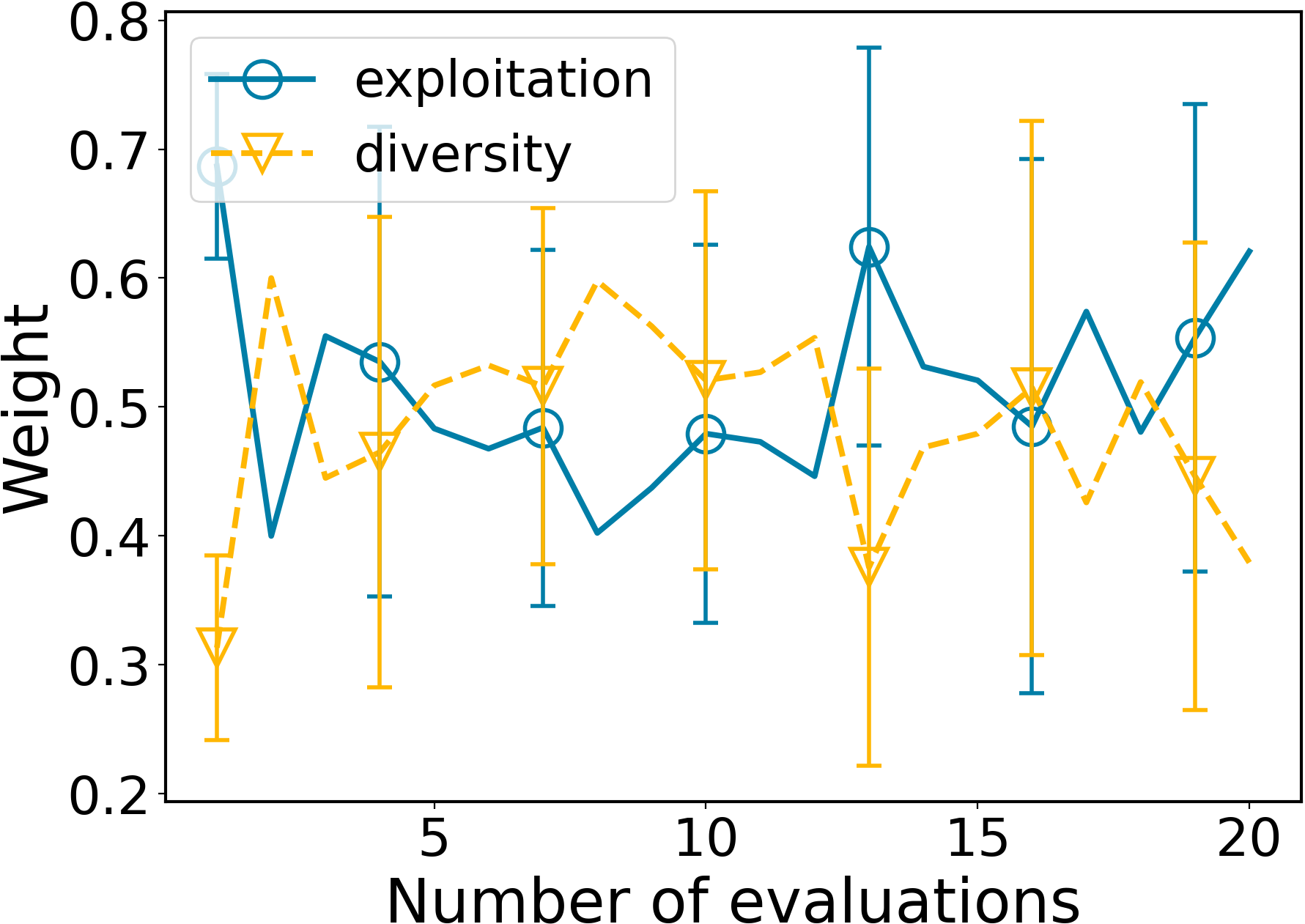}
        \caption{Diversity Metrics}
        \label{fig:rosen_div_metrics}
    \end{subfigure}
    \begin{subfigure}{0.3\textwidth}
        \centering
        \includegraphics[width=\linewidth]{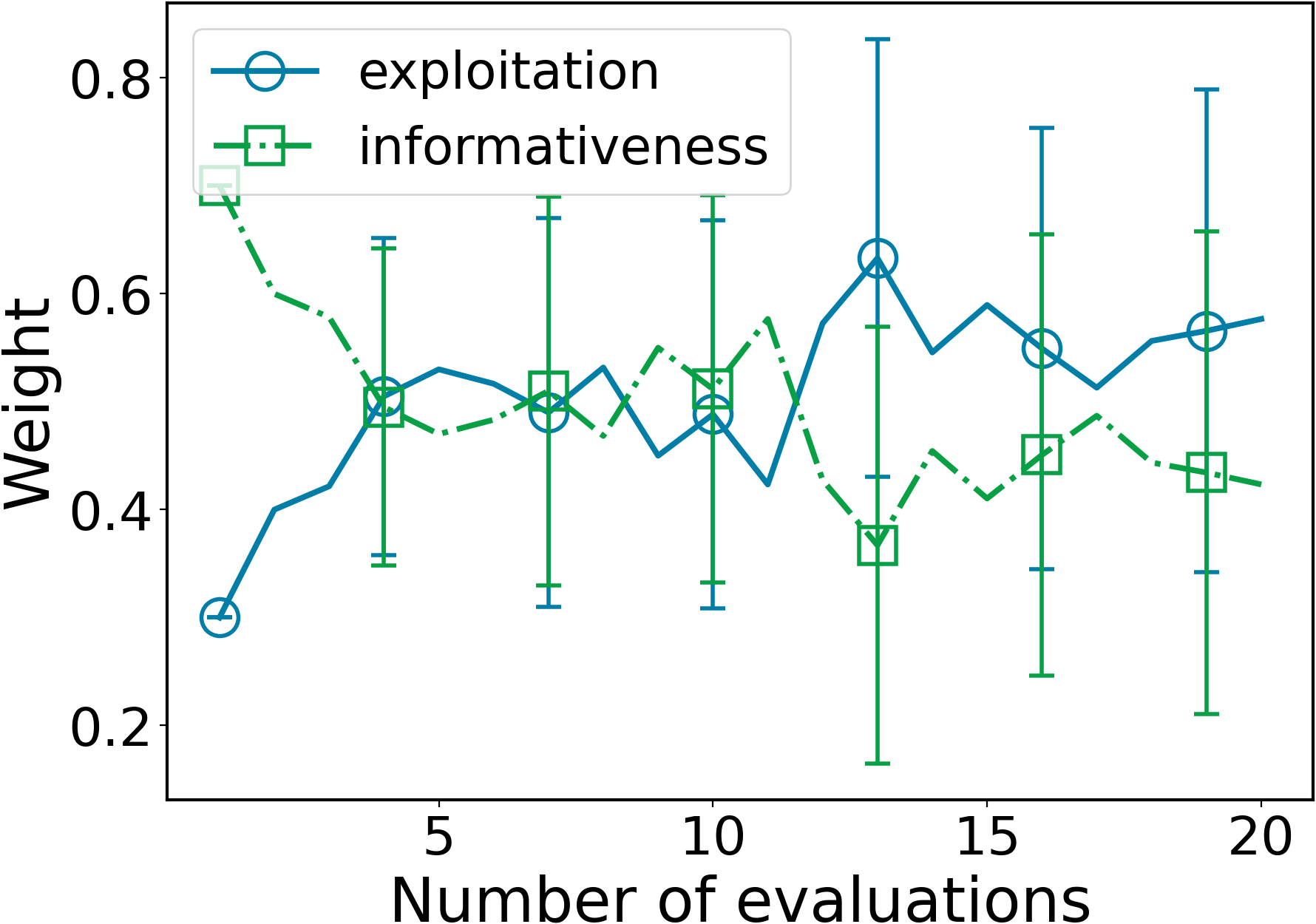}
        \caption{Informativeness Metrics}
        \label{fig:rosen_info_metrics}
    \end{subfigure}

    \caption{Performance and metric evolution for paired exploration criteria on the Rosenbrock function.}
    \label{fig:combination_results_rosen}
\end{figure}

For the robot pushing task (Figures~\ref{fig:robot_multi}–\ref{fig:robot_multi_metrics}), involving complex and discontinuous dynamics, the multi-agent LLM outperforms the single-agent variant and approaches the performance of EI and UCB. Early gains are driven by exploration, followed by steady improvement. Unlike Rosenbrock, the strategy maintains an exploitation-dominant policy complemented by persistent but moderate exploration, consistent with the demands of a nonconvex control landscape.
Paired-metric ablation studies provide further insights into these trends, by restricting the strategy agent to trade off exploitation against a single exploration criterion. 
On smooth analytical objectives such as Rosenbrock function, paired-metric configurations produced an improvement (Figures~\ref{fig:rosen_repr}-\ref{fig:rosen_info}), with the LLM systematically reallocating weight towards exploration in response to slow improvement (Figures~\ref{fig:rosen_repr_metrics}–\ref{fig:rosen_info_metrics}). In contrast, on robot pushing, performance degraded across metric pairings (Section~\ref{subsec:robot_pushing} - Figures~\ref{fig:robot_repr}-\ref{fig:robot_info}), and strategy weights remained stable (Section~\ref{subsec:robot_pushing} - Figures~\ref{fig:robot_repr_metrics}-\ref{fig:robot_info_metrics}), indicating that single pairs of metrics are not enough to navigate the complex landscape structure of the problem. Overall, these results highlight the importance of explicitly representing and controlling exploration–exploitation trade-offs when deploying LLM-based optimizers across diverse problem classes.

\subsection{Impact of Metric Selection on Strategy Evolution for Robot Pushing}\label{subsec:robot_pushing}

\begin{figure}[h!]
    \centering
    \begin{subfigure}{0.28\textwidth}
        \centering
        \includegraphics[width=\linewidth]{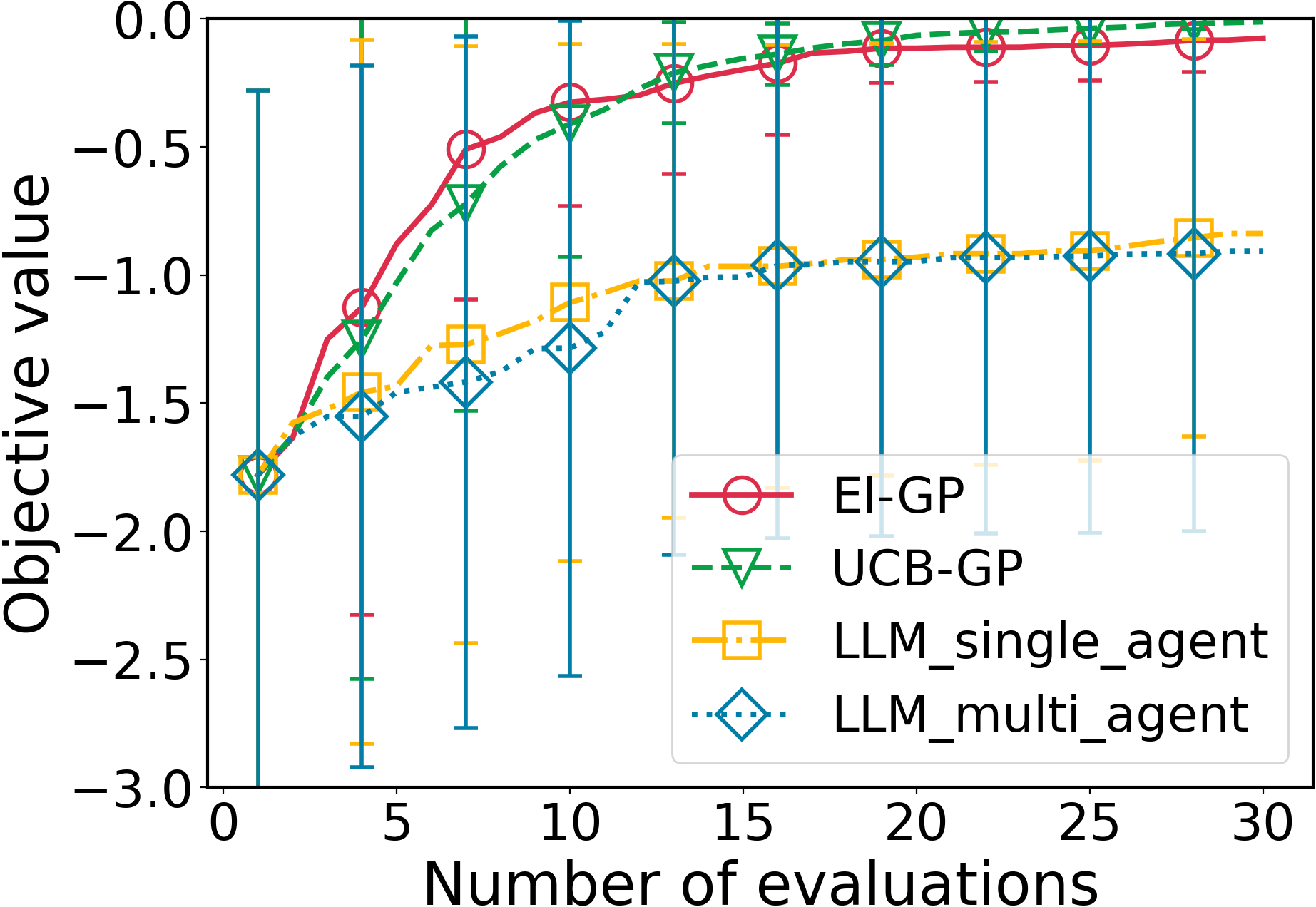}
        \caption{Representativeness Results}
        \label{fig:robot_repr}
    \end{subfigure}
    \begin{subfigure}{0.28\textwidth}
        \centering
        \includegraphics[width=\linewidth]{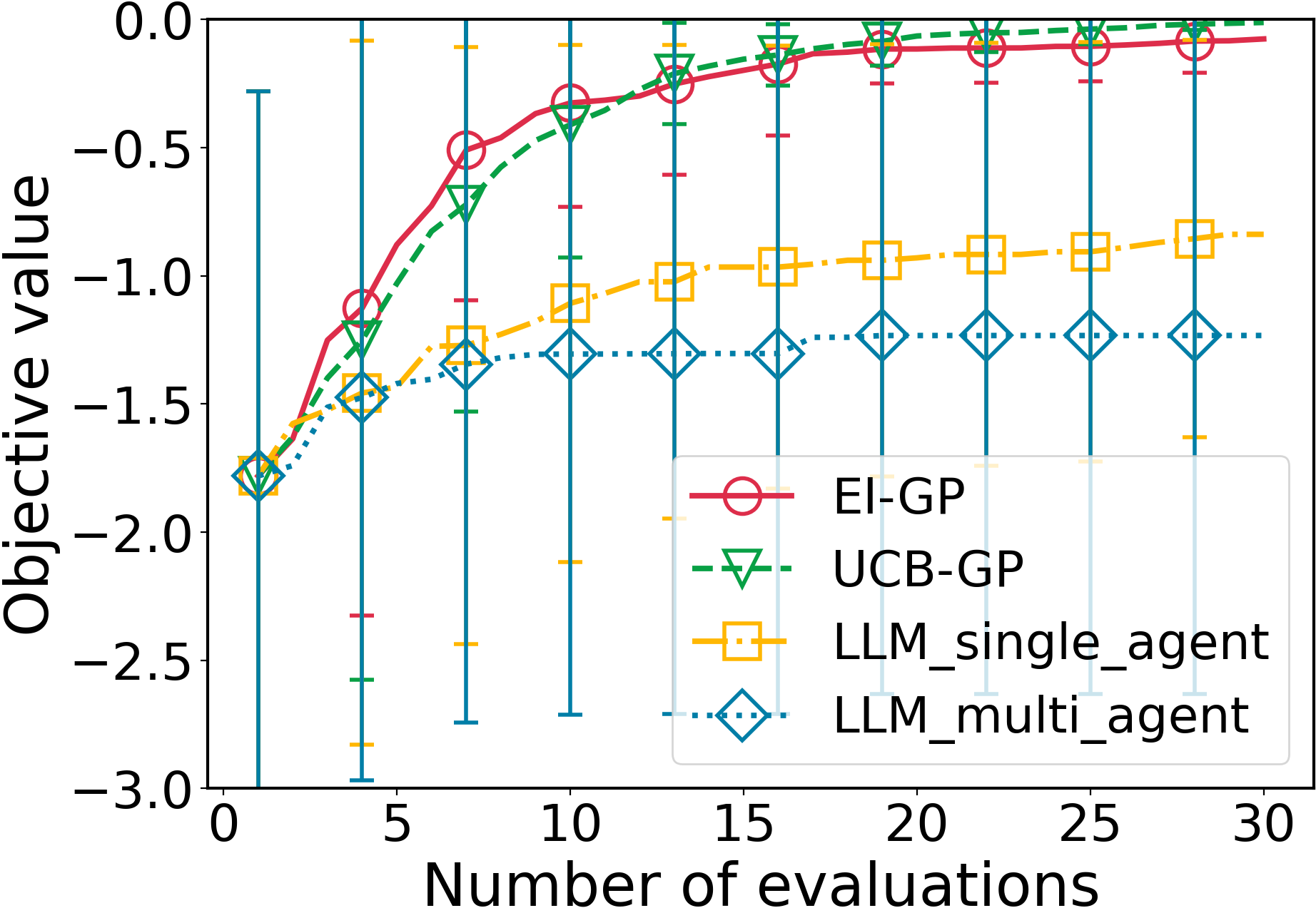}
        \caption{Diversity Results}
        \label{fig:robot_div}
    \end{subfigure}
    \begin{subfigure}{0.28\textwidth}
        \centering
        \includegraphics[width=\linewidth]{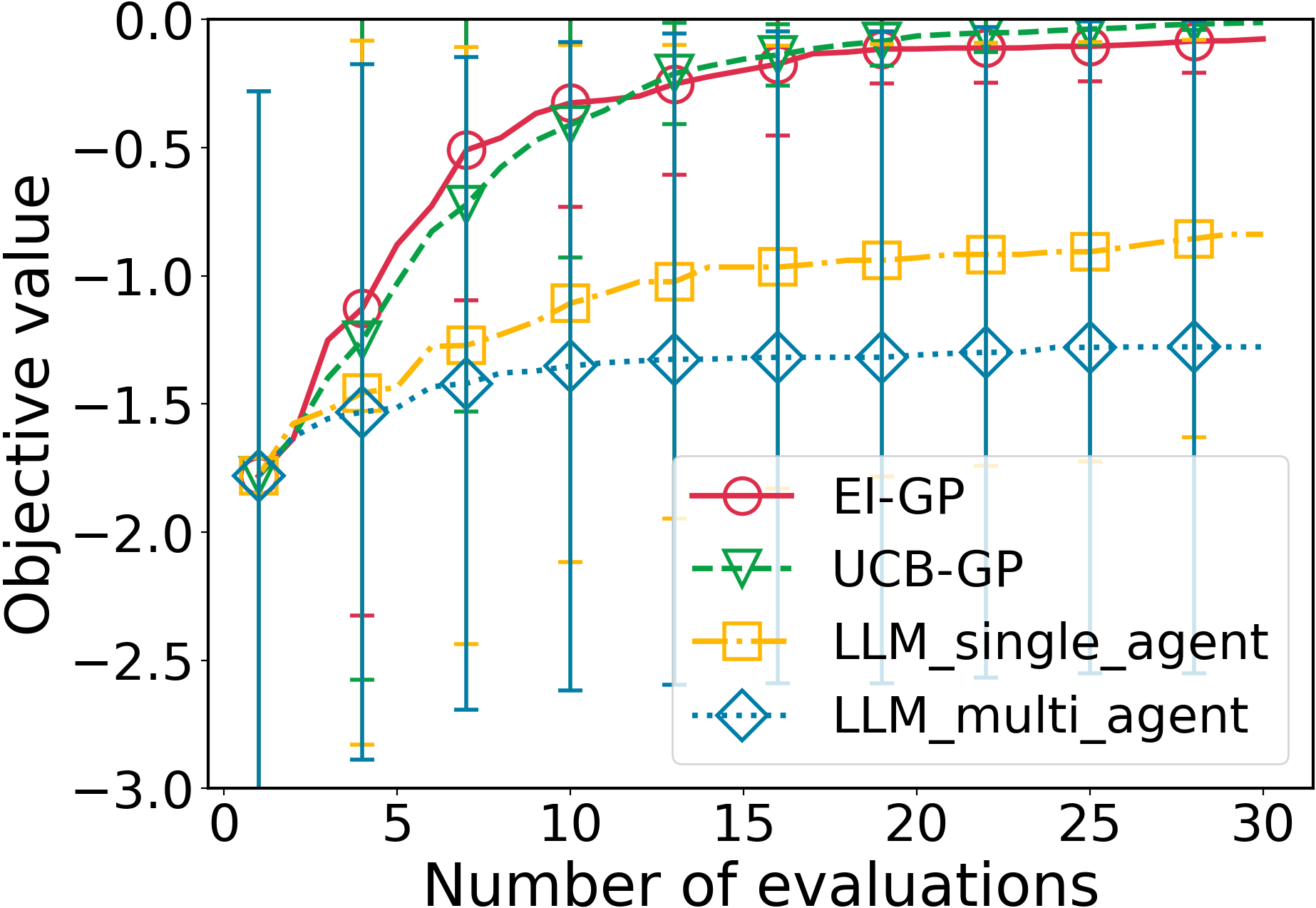}
        \caption{Informativeness Results}
        \label{fig:robot_info}
    \end{subfigure}

    \begin{subfigure}{0.28\textwidth}
        \centering
        \includegraphics[width=\linewidth]{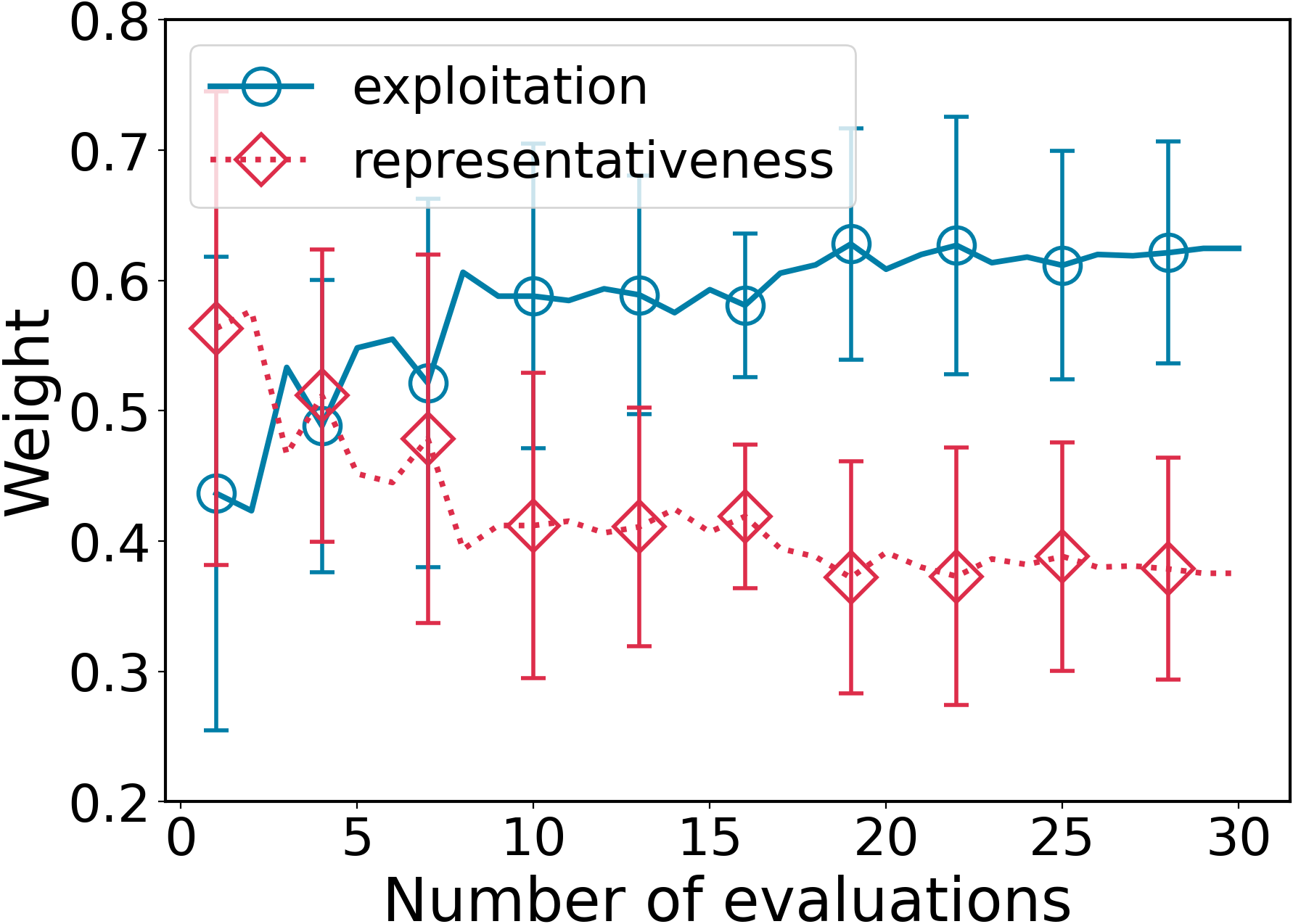}
        \caption{Representativeness Metrics}
        \label{fig:robot_repr_metrics}
    \end{subfigure}
    \begin{subfigure}{0.28\textwidth}
        \centering
        \includegraphics[width=\linewidth]{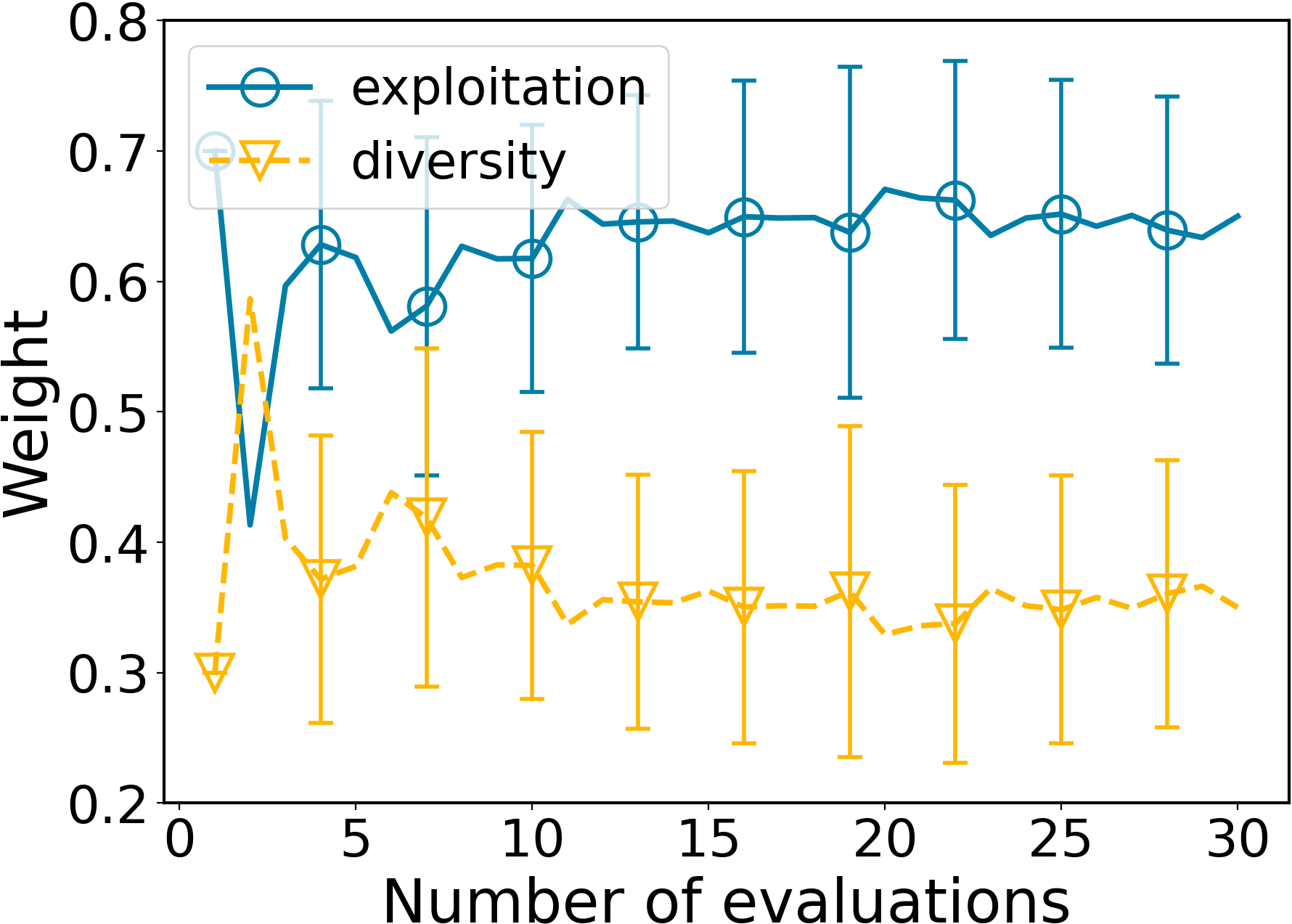}
        \caption{Diversity Metrics}
        \label{fig:robot_div_metrics}
    \end{subfigure}
    \begin{subfigure}{0.28\textwidth}
        \centering
        \includegraphics[width=\linewidth]{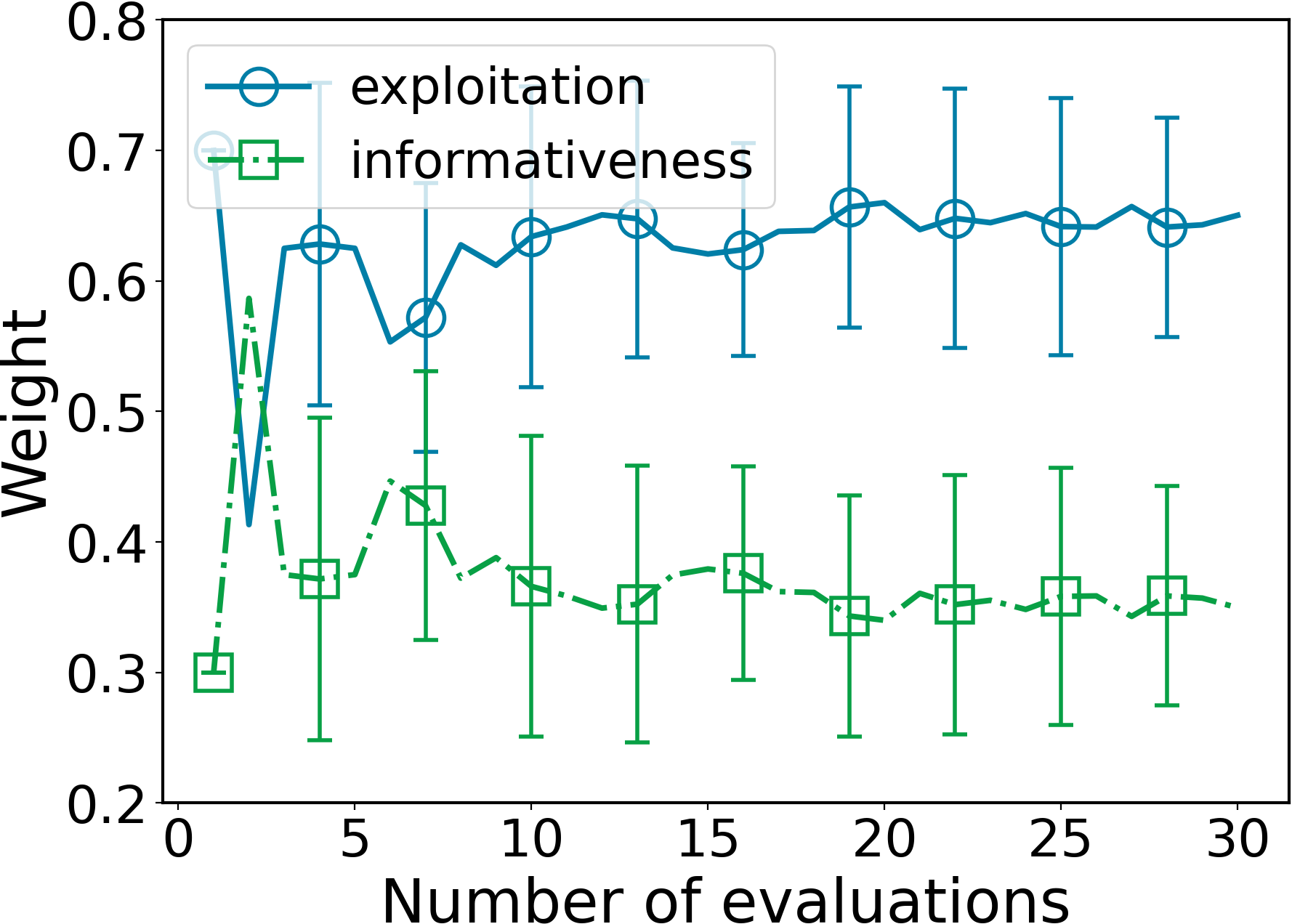}
        \caption{Informativeness Metrics}
        \label{fig:robot_info_metrics}
    \end{subfigure}

    \caption{Performance and metric evolution for paired exploration criteria on the robot pushing task.}
    \label{fig:combination_results_robot}
\end{figure}

As displayed in Figures~\ref{fig:robot_repr}-\ref{fig:robot_info}, coupling \textit{Exploitation} with a single auxiliary metric did not produce the same performance. The use of \textit{Representativeness}, \textit{Diversity}, and \textit{Informativeness} alone did not improve over the Single Agent baseline.
As shown in Figures~\ref{fig:robot_repr_metrics}–\ref{fig:robot_info_metrics} the weight trajectories exhibit premature stagnation across metric pairs, suggesting couple of metrics alone are not sufficient to navigate the complex problem space.

\section{Conclusion}\label{sec:conclusion}

This work analyzes how LLMs mediate exploration–exploitation trade-offs in black-box optimization by explicitly externalizing search policies through interpretable metrics. Treating LLMs as policy mediators rather than surrogates or end-to-end optimizers provides a principled lens for examining how competing notions of exploration, such as diversity, informativeness, and representativeness, are adaptively prioritized across problem contexts. Empirically, separating strategy selection from candidate generation stabilizes search dynamics and improves performance in weakly structured settings, while simultaneously exposing systematic limitations in smooth landscapes that demand fine-grained numerical exploitation. Notably, LLM-mediated strategies do not recover classical acquisition functions; instead, they construct dynamic, context-sensitive policies that react to perceived stagnation and uncertainty, with effects that can be either beneficial or detrimental depending on landscape structure. Paired-metric ablations show that failures on complex objectives arise not from a specific exploration criterion, but from a tendency to over-diversify when progress slows, highlighting the need for explicit control of the exploration–exploitation balance.

Several directions emerge for future work. Incorporating lightweight numerical feedback or local refinement mechanisms may alleviate deficiencies in smooth regimes without sacrificing interpretability. Extending the policy mediation framework to richer or adaptive metric sets, including hierarchical or discovered strategies, could improve robustness across heterogeneous problem classes. Finally, integrating formal uncertainty or regret estimates into policy mediation, without reverting to full probabilistic surrogates, offers a promising path toward reconciling LLM-based reasoning with classical optimization theory, advancing interpretable and controllable LLM-driven optimization systems.

\section*{Acknowledgments}

This work was supported in part by the U.S. National Science Foundation under Grant No. CNS-2320261, Research Infrastructure: MRI: Track 2 Acquisition of Data Observation and Computation Collaboratory (DOCC) (PI: Renambot). 
The authors acknowledge the use of computational resources provided by the University of Illinois Chicago under this award.

\bibliography{bibliography}

\end{document}